# Hybrid Medical Image Classification Using Association Rule Mining with Decision Tree Algorithm

P. Rajendran, M.Madheswaran

**Abstract**— The main focus of image mining in the proposed method is concerned with the classification of brain tumor in the CT scan brain images. The major steps involved in the system are: pre-processing, feature extraction, association rule mining and hybrid classifier. The pre-processing step has been done using the median filtering process and edge features have been extracted using canny edge detection technique. The two image mining approaches with a hybrid manner have been proposed in this paper. The frequent patterns from the CT scan images are generated by frequent pattern tree (FP-Tree) algorithm that mines the association rules. The decision tree method has been used to classify the medical images for diagnosis. This system enhances the classification process to be more accurate. The hybrid method improves the efficiency of the proposed method than the traditional image mining methods. The experimental result on prediagnosed database of brain images showed 97% sensitivity and 95% accuracy respectively. The physicians can make use of this accurate decision tree classification phase for classifying the brain images into normal, benign and malignant for effective medical diagnosis.

**Index Terms—** Image minig, Data mining, Association Rule Mining, Decision Tree Classification, Watershed algorithm, FP-Tree algorithm,Medical Imaging.

## 1 INTRODUCTION

A brain tumor is an abnormal growth of cells within the brain or inside the skull, which can be cancerous (malign) or non-cancerous (benign). It is defined as any intracranial tumor created by abnormal and uncontrolled cell division. This type of brain tumor constitutes one of the most frequent causes of death among the human being in the world. Detection of tumor in the earliest stage is the key for its successful treatment. One of the famous method used recently for the screening procedure from the patients include CT-Scan (Computerized Tomography Scan) brain images [1][2]. From the CT-Scan brain images the radiologist will be able to diagnose the abnormalities in the tissues. Even though some 10 to 30% of the tumor cells will not be able to be diagnose correctly. Hence the Computer Aided Diagnosis (CAD) system will assist the physicians as a “Second option” in clearly diagnosing the cancerous cell in CT-Scan brain images [2].

The proposed method classifies the CT-Scan brain images into three types: Normal, benign, and malignant. The normal images depict the cells of healthy patients, benign cells are like cancerous cells but not originally cancerous and third type is malignant cells that depict the original steps for classifying the CT-Scan brain images into the cancerous cells. The proposed method consists of various above mentioned three types of cells they are: preprocessing, feature extraction, rule generation, classification and diagnosis.

In this paper image mining concepts have been used. It deals with the implicit knowledge extraction, image data relationship and other patterns which are not explicitly stored in the images. This technique is an extension of data mining to image domain. It is an inter disciplinary field that combines techniques like computer vision, image processing, data mining, machine learning, data base and artificial intelligence [5]. The objective of the mining is to generate all significant patterns without prior knowledge of the patterns [6]. Rule mining has been applied to large image data bases [7]. Mining has been done based on the combined collections of images and it is associated data. The essential component in image mining is the identification of similar objects in different images [8].

The Watershed morphological transformation of images has been explained for segmentation and removal of inconsistent data from the image [12], [13]. Feature extraction is one of the most important steps in image extraction [4] [20]. Association rule mining has been used in most of the research for finding the rules for diagnosis in large and small databases [3], [9], [10], [15]. In this proposed method FP-Tree method has been used to find the frequent pattern for building the association rule [10], [16],[17],[18].

• P.Rajendran is with Department of CSE, K. S. Rangasamy College of Technology, Tiruchengode - 637215, Tamilnadu, India.

• M.Madheswaran is with Center for Advance Research, Department of Electronics and Communication Engineering, Muthayammal Engineering College, Rasipuram-637 408, Tamilnadu, India.

The classification and decision tree construction of extracted images can be a very important method for the diagnosis. The frequent-pattern growth algorithm is standard ARM algorithm, which is efficient for mining large dataset s with frequent patterns. Its efficiency lies in the compact and complete way it represents the entire set of transactions and patterns in a tree-like form, which eliminates the need to generate candidate items sets [11], [14], [24], [25].

The following sections describe the various techniques that have been implemented in the proposed method. Section 2 describes the system description, section 3 describes the construction of association rule, section 4 explains about buliding the Hybrid classifier, section 5 details about the Results and discussions, section 6 conclusion.

## 2 SYSTEM DESCRIPTION

The proposed method has been divided into two main phases: the training phase and the test phase. Various techniques followed in these phases are, pre-processing, feature extraction, rule generation classification and Diagnosis. The pre-processing and feature extraction technique are common for both training and test phase. The overview of the proposed system is given in the fig. 1.

In the training and test phases the acquired images have been taken for the preprocessing and feature extraction process. The preprocessing has been done by using the median filtering with morphological opening process. Edges are segmented using canny edge detection technique. The regions are extracted in the feature extraction phase. These features are stored in the transactional database. The FP-tree method generates the maximum frequent items that are stored in the transaction database. Association rule can be constructed using maximum frequent itemset that are generated from the FP tree algorithm [16], [17].The association rules based classifications have been made with the help of decision tree classification. This hybrid approach has been used to classify the CT-Scan brain images into normal, benign and abnormal.

### 2.1 Pre-processing and edge detection

The prime objective of the preprocessing is to improve the image data quality by suppressing undesired distortions (or) enhancing the required image features for further processing. The irrelevant data present in the image has been eliminated using the pre-processing technique. The pre-processing technique eliminates the incomplete, noisy and inconsistent data from the image in the training and test phase. In order to improve the quality of images taken from the CT-scan brain images and to make the feature extraction phase more reliable, pre-processing is necessary.

During the digitization process, noise could be introduced that needs to be reduced by applying median filtering techniques. Normalisation histogram of the image provides the contrast information and overall intensity distribution. An intensity normalization procedure is carried out by computing an average intensity histogram on the basis of all the images, detecting its peak value, and then adjusting each single image so that its histogram peak aligns to the average histogram. Thus the brightness of all pixels in the image is readjusted according to the scale of the gray level of the images ranged from 0 to 255. The normalized image is further smoothed using 2-D median filter.

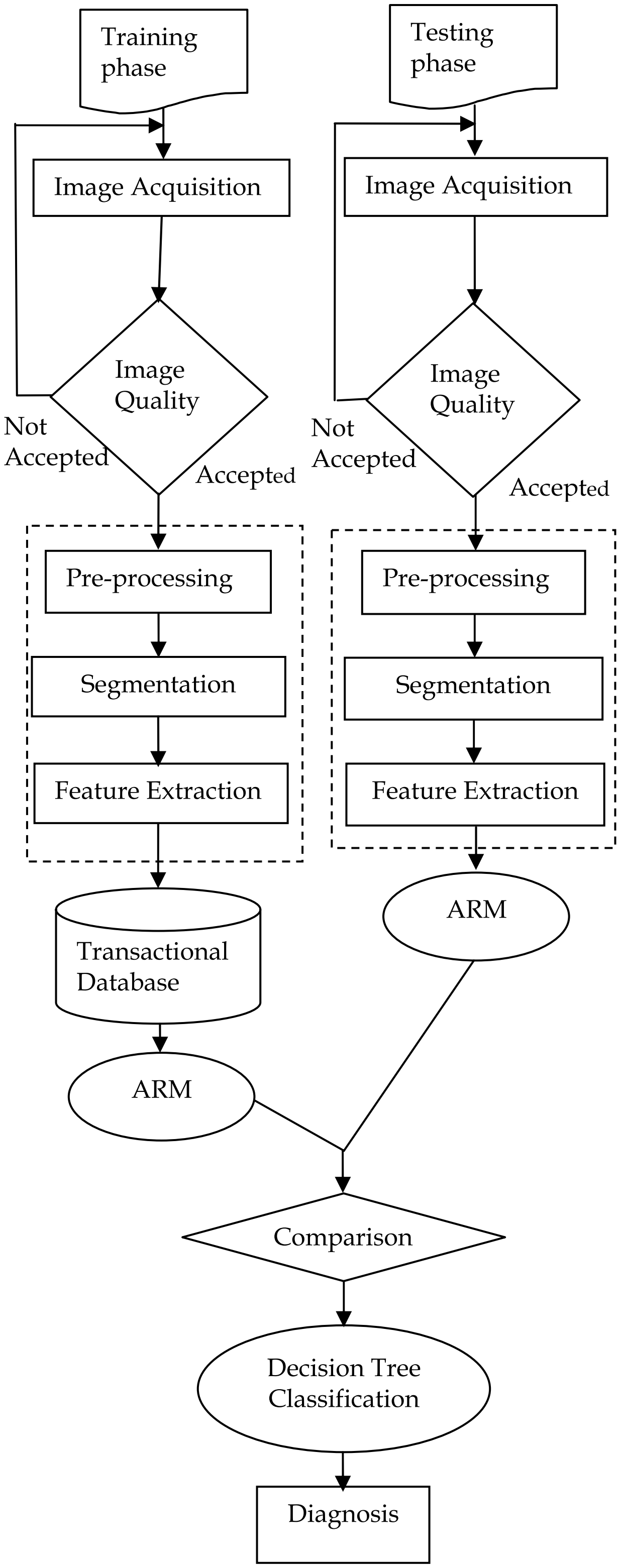

**Fig. 1 Overview of the proposed system**

### 2.1.1 Median Filtering

Histogram equalization [1, 11] helps to improve the contrast of the image without affecting the structure. While this technique gives the output image with increased contrast, it also causes an increase in the noise within the CT image. In this step, the images are smoothed horizontally to overcome the difficulties using the smoothing filter. 2D median filter (3×3 square filter) has been used prior to edge detection to reduce the amount of noise in the stack of 2D images [1, 11]. This is preferred since it is a very simple smoothing filter but at the same time preserves the edges.

### 2.1.2 Morphological opening

Opening operation can be used to smoothes the contour of an object, breaks naro isthumuses and eliminates thin protrusions. The opening of set A by structuring element B, denoted as $A \circ B$ is defined as

$$A \circ B = (A \Theta B) \oplus B$$

Thus, opening A by B is the erosion of A by B, followed by a dilation of the result by B [29], [30].

### 2.1.3 Edge Detection

The edge detection method called the canny edge detection with the threshold value of 0 has been used. Edge feature along with the color feature gives good efficiency. The combination of edge and color feature describes the boundaries and inner regions of tumor cells. The features that are extracted have been stored in the transactional database to from the association rule.

**Edge detection Procedure**

Edge detection technique can be used significantly to reduce the amount of data in an image and at the same time structural properties of an image can also be preserved for further analysis. In this proposed method, canny edge detection technique has been used with the following criteria.

- **Detection**: The probability of detecting real edge points should be maximized while the probability of falsely detecting non-edge points should be minimized. This corresponds to maximizing the signal-to-noise ratio.
- **Localization**: The detected edges should be as close as possible to the real edges.
- **Number of responses**: One real edge should not result in more than one detected edge.

The canny edge detector was designed to be the optimal edge detector. The two main goals are to minimize the error rate (i.e. the amount of false positives and false negatives) and to provide localized edges. The Canny edge detector simplifies the process by using convolutions with Gaussians.

The first step is to find the gradient in the x and y directions ($g_x$ and $g_y$) by smoothing the image and taking the derivative. If a two-dimensional Gaussian $G_{xy}$ is convolved with the image I for smoothing, the operation is given by equation 1.

$$g_x = \partial / \partial x(I * G_{xy}) \text{ And } g_y = \partial / \partial y(I * G_{xy}) \qquad (1)$$

The * operator denotes convolution. The form above is not the most computationally efficient method of computing the gradient. Using the fact that the Gaussian is separable, a more efficient form can be found as shown in equation 2.

$$g_x = \partial / \partial x(I * G_{xy}) = I * \partial / \partial x(G_{xy}) = I * \partial / \partial x(G_x * G_y) = I * \partial / \partial x(G_x) * G_y \qquad (2)$$

Likewise, the gradient in the y direction is calculated in equation 3.

$$g_y = I * G_x * \partial / \partial y(G_y) \qquad (3)$$

The $g_x$ and $g_y$ give the gradient estimation in the x and y directions, respectively. Let $g_x(x, y)$ be the gradient in the x direction at (x, y), and $g_y(x, y)$ be the gradient in the y direction. The magnitude of the gradient g(x, y) is given in the equation 4.

$$g(x, y) = \sqrt{g_x(x, y)^2 + g_y(x, y)^2} \qquad (4)$$

If processing time is a factor, the above equation can be approximated in the equation 5.

$$g(x, y) = |g_x(x, y)| + |g_y(x, y)| \qquad (5)$$

The direction of the maximum slope is the angle formed by the vector addition of the two gradients in the x and y directions. This angle $\theta(x, y)$ is found from the equation 6.

$$\theta(x, y) = \tan^{-1}\left[\frac{g_y(x, y)}{g_x(x, y)}\right] \qquad (6)$$

The fig. 2 shows the angles that will be returned from the arctangent and the pixels to which these angles correspond.

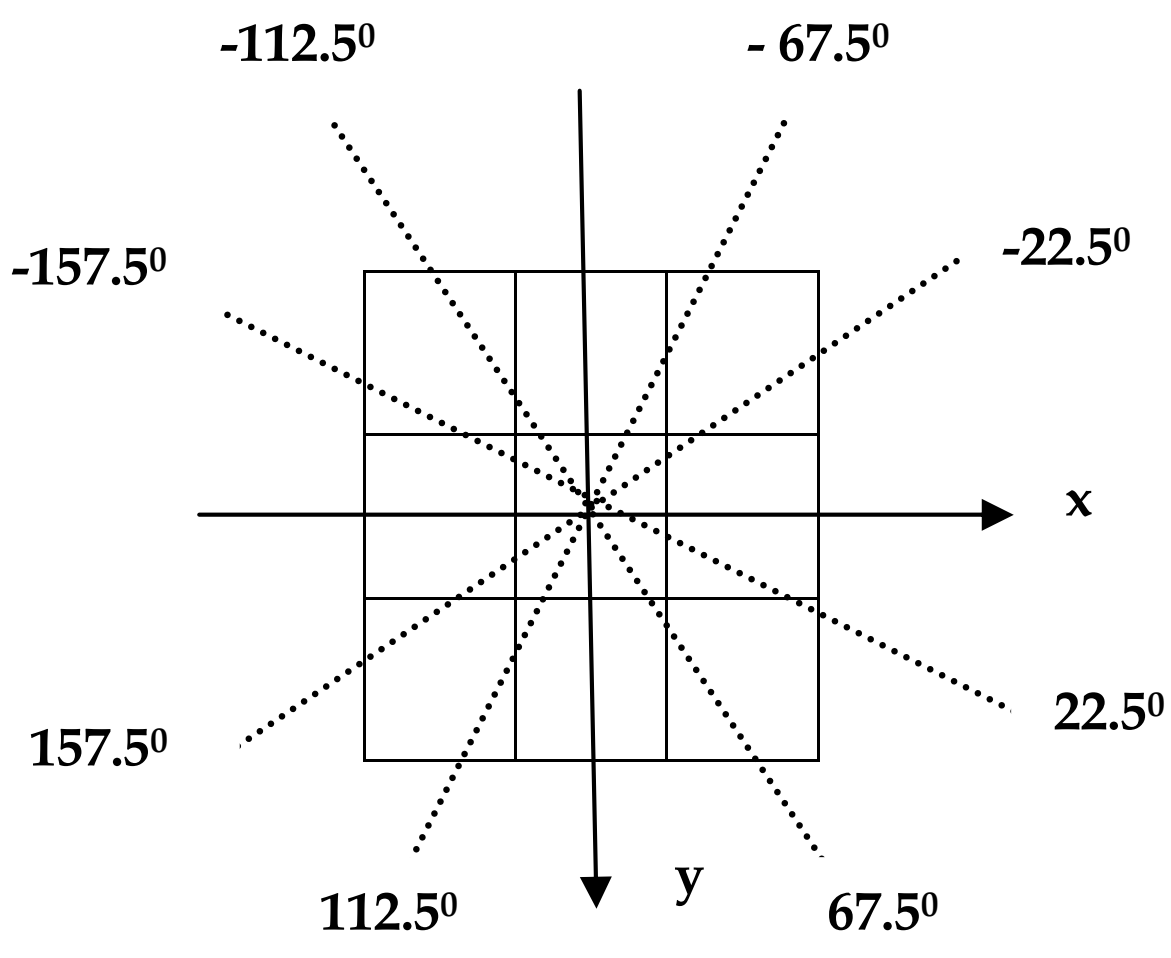

Fig. 2 arctangent and the pixels towhich these angles correspond

For the non-maximum suppression, the neighboring pixels in the direction of the gradient are needed. To find these pixels, add 180° to $\theta$, if $\theta$ is negative and use the rules are given in the table 1.

TABLE 1 IF $\theta$ IS NEGATIVE AND USE THE RULES

| Theta Values (in degree) | Direction (in degree) | Pixels |
|---|---|---|
| $\theta \leq 22.5$ *or* $\theta > 157.5$ | $\theta = 0$ | $(x-1, y)$ $(x+1, y)$ |
| $22.5 < \theta \leq 67.5$ | $\theta = 45$ | $(x-1, y-1)$ $(x+1, y+1)$ |
| $67.5 < \theta \leq 112.5$ | $\theta = 90$ | $(x-1, y+1)$ $(x+1, y-1)$ |
| $112.5 < \theta \leq 157.5$ | $\theta = 135$ | $(x, y+1)$ $(x, y-1)$ |

The next step is to perform non-maximum suppression on the image. The algorithm is to check every pixel in the image. If either of the two neighboring pixels in the direction of the gradient are greater in magnitude than the current pixel then I(x, y)=0, else I(x, y)=g(x, y). The effect of this operation is to create one-pixel thick lines along the local maxima of the gradient.

Once the non-maximum suppression has been performed on the image, the edge detector must differentiate between lines that are edges and lines which are due to noise. To make this distinction, the canny edge detector uses hysteresis thresholding. This method of thresholding involves a high and a low threshold. All the pixels that have a gradient value above the high threshold will be retained. Similarly, all the pixels that have a gradient value below the low threshold will be rejected. The pixels with gradient values in between the high and low thresholds are retained only if the pixel is connected to another pixel with a gradient value above the high threshold. The output of the hysteresis thre thresholding produces the final image.

The Chamfer distance is a method of finding the distance to the edges in the image. The Manhattan distance is used in this assignment. The Manhattan distance D for two points $p = (p_x, p_y)$ and $q = (q_x, q_y)$ is given in equation 7.

$$D(p,q) = |q_x - p_x| + |q_y - p_y| \tag{7}$$

The Chamfer distance is used in this assignment to find the correct placement of a template image on the original image. This is accomplished by computing a probability map of the sum of the distances to the edges.

The given input image is segmented in to the number of objects using canny edge detection method. For each object the texture features have been calculated and stored in the transactional database.

## 3 ASSOCIATION RULE MINING

One of the most important Knowledgw Distrubuted Database research issues is association rules. ARM task is to discover the hidden association relationship between the different itemsets in transaction database [3], [9], [14], [15].

### 3.1 Frequent Itemsets and Maximum Frequent Itemsets

Let I={$i_1$,$i_2$,….$i_m$} be a set of m distinct items. A transaction T is defined as any subset of items in I. A transaction database T is said to support an itemset $x \subseteq I$ if it contains all items of x. The fraction of the transaction in D that support x is called support value is above some user defined minimum support threshold then the itemset is frequent, otherwise it is infrequent. Maximum frequent itemsets have been denoted as F if all superset of frequent itemsets is infrequent itemsets. The maximum frequent itemsets that are discovered have been stored in the maximum frequent itemsets.

Maximum frequent candidate set [14], which is the smallest itemsets, it includes all current frequent itemsets known, but it does not include any infrequent itemsets. The identification of maximum frequent itemsets in earlier stage can reduce the number of candidate itemsets genetrated. So that it will reduce the CPU and I/O time. In case if the maximum frequent itemsets discovered are long, then the performance of the algorithm will be excellent, so the issue of discovering frequent itemsets can be converted into the issue of discovering maximum frequent itemsets.

All the maximum frequent itemsets identified has been stored in the transactional database. The information stored in the transactional table is given in the form of tree structure is shown in the figure 3. The encoded information has been given clearly in the form of tree structure in figure 4.

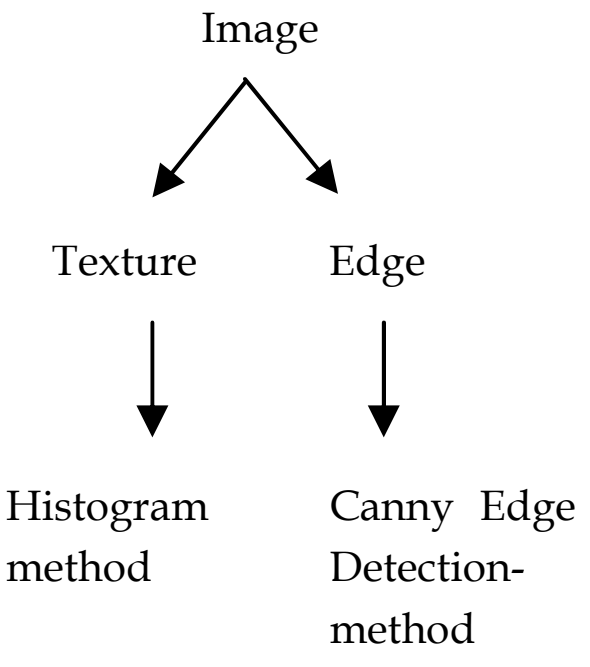

Fig. 3 Hierarchical tree

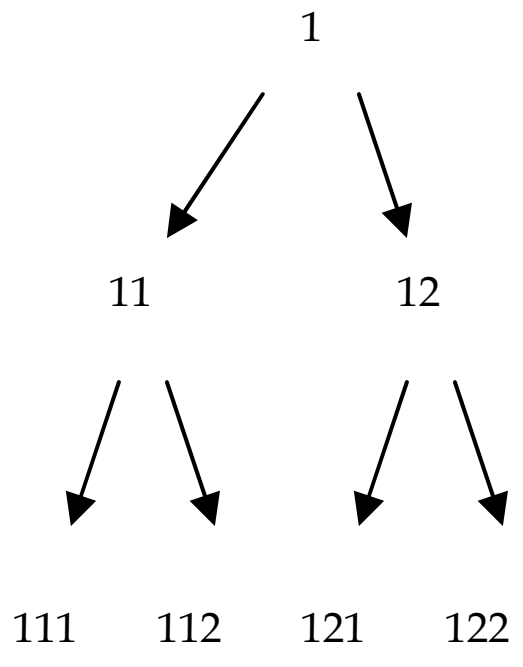

Fig. 4 Encoded Hierarchical Tree

### 3.2 Frequent Pattern Tree

The special prefix tree called FP-Tree consists of a frequent item header table. The root is labeled as "NULL" and a set of the item prefix subtree are called the children. There are four fields defined for each item prefix subtree, they are node-name, node-count, node-link node-count and node-parent [16], [17]. There are also three fields are included for each entry of the frequent item header table, those are item-name, item-sup, item-head. The pointer pointing to the first node in the FP-Tree carrying node name is item-head algorithm. The FP-Tree construction steps includes,

**First TDB Scan**:

Let L be the set of frequent items and support value of each frequent item have been collected.

(i) Scan TDB again:

Let L be the set of frequent items and support value of each frequent item have been collected.

(ii) Scan TDB again:

All frequent items in every transaction has been selected and sorted according to the order of list of frequent items.

From the corresponding itemsets by combining the ordered frequent items in every transaction and insert them to FP-Tree respectively [17].

### 3.3 Maximum Frequent itemsets mining

Definition 1: If an itemset is infrequent, all its superset must be infrequent.

Definition 2: If an itemset is frequent, all its subsets must be frequent.

Definition 3: If A={a1,a2,…..am-1} is an infrequent itemset, and ai={aj/ai ∈ A, j ≠ i} may be frequent itemset.

The top-down search algorithm uses the concept os definition 2, which begins from the only itemsets, it reduces some candidate itemsets in every pass. According to the definition 2,if K-itemsets is infrequent in K pass, then check all (K-1) itemsets in next pass consequently it can discover all maximum frequent itemsets. The definition 4.1 can be used in the situation where maximum frequent itemsets are short. Hence in the proposed system, top-down search tactic is used which uses the maximum frequent candidate set to store the information. The support value of each maximum frequent itemsets has been counted respectively.

TABLE 2 TRANSACTION TABLE

| TID | TRANSACTION |
|---|---|
| 001 | 111,121,211,221 |
| 002 | 11,211,222,323 |
| 003 | 112,122,221,421 |
| 004 | 111,121,421 |
| 005 | 111,122,211,221,413 |
| 006 | 211,323,524,413 |
| 007 | 323,524,713 |

Maximum Frequent Itemset Finding Algorithm:

**Input**: FP-tree generated from transaction database; frequent item header table and the minimum threshold support (minsup) defined by user; lest of frequent itemsets LF = {1,2.3…..k}

**Output**: Maximum frequent itemsets (MFI)

(1) For ( i=1;i<=max_level; i++) // Where I is a variant which represents corresponding level in the hierarchical association rules.
(2) MFI i = ϕ ;
(3) MFC = FL = {1,2,3,….K}; // MFC is the maximum frequent candidate sets in Transactional database.
(4) While ( MFC ≠ ϕ ) do begin
(5) for (j=k; j>0; j--) do begin
(6) MFC j= { c/ C ∈ MFC and item is the last item in c };
(7) MFC = MFC – MFCj;
(8) Call Compute Count ( FP_tree, frequent item header table , MFC);
(9) for all n ∈ MFCj do begin
(10) if n.support >= minsup then
(11) MFIi= MFIi ∪ n
(12) Else
(13) For all item m ∈ n do

(14) If n = {m} is not subset of each element in MFIi and MFC then
(15) MFC = MFC ∪ {n-{m}};
(16) End
(17) End
(18) MFI = MFI ∪ MFIi
(19) End

*Theorem 1*: Let be transaction database TDB and a support threshold value mins up, have been given. The support value of every frequent itemsets in the transaction database can be desired from the transaction database's FP tree. Generally, based on the FP – tree algorithm, for each transaction in the transaction database TDB, the projection of the frequent item has been mapped to one path in the FP- tree Given a frequent itemsets A={$a_1$,$a_2$,…..$a_n$} and an item $a_i$(1<= i <= n) has ordered by support descending in TDB. According to the node link of item an is FP-tree, and according to the node-parent field of an item an , discover all path that include item an as the last item. For all the path in the FP-tree constructed that includes the nodes from root to node V has been racked as an, the support value of node V indicates transaction numbers that includes path P. If $a_1$, $a_2$,…..$a_n$ are all is the path P and it shows that path p includes A, so the number of transaction which include p is node-count v . Hence, the support value of frequent itemsets accumulated by the method given above, provide the total number is support of X.

The Maximum frequent itemset finding algorithm employs the function Compute Count to count the support value of every itemsets is MFCi which is based on theorem 1, hence all items from the itemsets is MFCi have been ordered by support-descending consequently, For the confirmation of the number of path that includes the exact itemsets, the above given algorithm is convenient. The FP-tree based on the transactional database is illustrated in the figure 5.

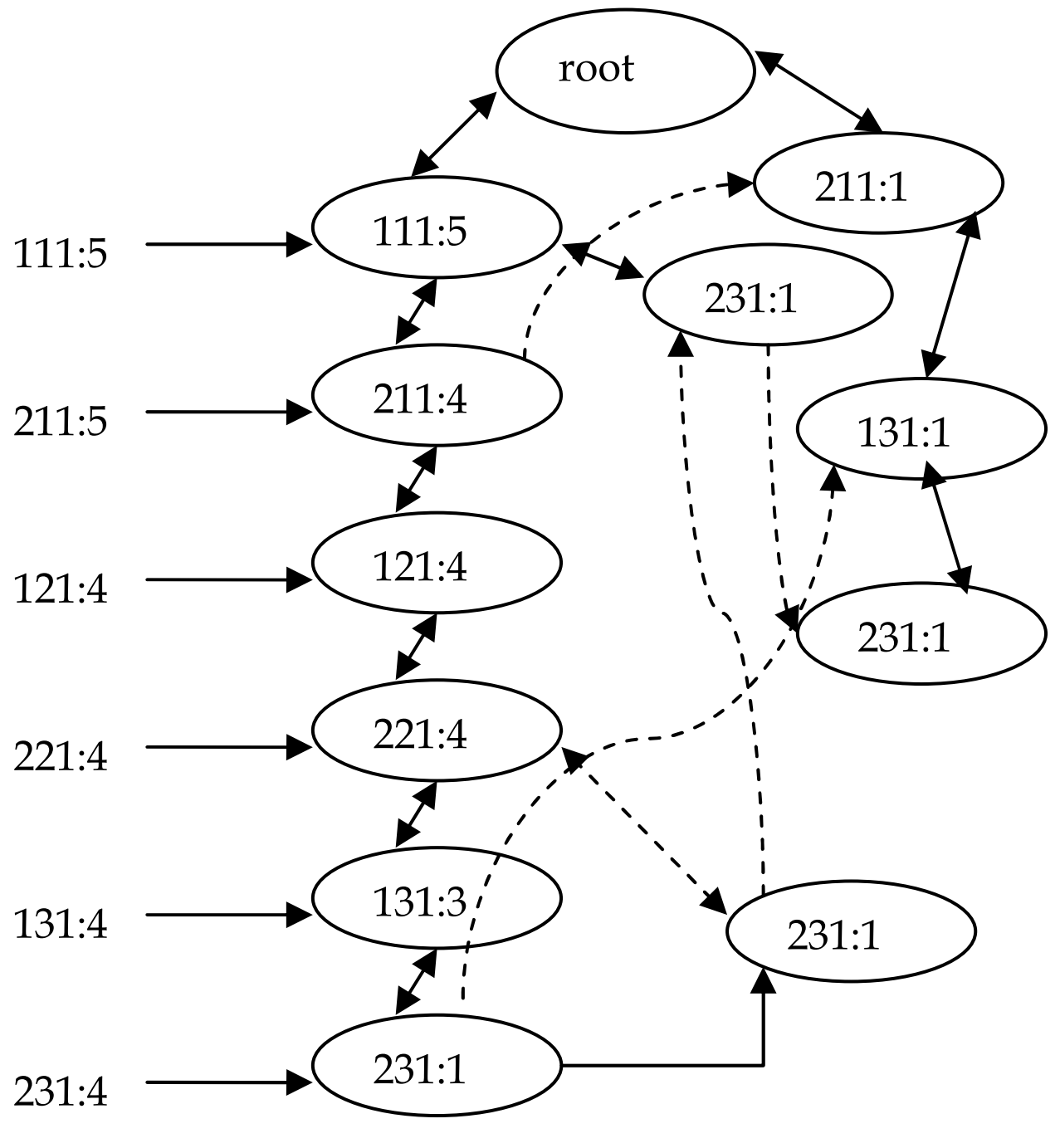

Fig. 5 FP-Tree based on transaction table

# 4 BUILDING THE HYBRID CLASSIFIER

## Hybrid Association Rule with Decision Tree Classification

Decision tree based classification methods are widely used in data mining and decision support application. In the proposed system,decisions that have to be made by the physicians whether the maximum frequent itemset that are found in the FP-tree method has been compared with the maximum frequent item of the test images [23], [24].

In the proposed system, the decision tree classification with association rule classification method provides a better option for the phycisians classify the benign and malignant images. It is done by comparing the maximum frequent items generated by the association rules in the training image have been compared with the maximum frequent items of the test image and hence the diagnosis can be made easily. The Hybrid Association Rule Classifier (HARC) algorithm used in the proposed system, which classifies the rules generated by the association rules in to normal, benign or malignant. HARC builds decision trees from a set of training data, using the concept of information entropy. The training data is a set S = $s_1$, $s_2$,... $s_n$ of already classified samples. Each sample si = $x_1$, $x_2$,... $x_m$ is a vector where $x_1$, $x_2$,... $x_m$ represent attributes or features of the sample. The training data is augmented with a vector C = $c_1$, $c_2$,...$c_m$ where $c_1$, $c_2$ represent the class to which each sample belongs.

At each node of the tree, HARC chooses one rule of the data that most effectively splits its set of samples into subsets enriched in one class or the other. Its criterion is the normalized information gain (difference in entropy) that results from choosing an attribute for splitting the data. The attribute with the highest normalized information gain is chosen to make the decision. The HARC algorithm then recurses on the smaller sublists. The HARC algorithm is explained below and the tree generated by the HARC algorithm is given in the figure 6.

Algorithm: HARC

**Input:** (R: a set of non target Keywords, C: a the keyword, D: a training set )
**Output**: return a decision tree;

(1) begin
(2) if D is empty, return a single node with value Failure;
(3) if D consists of records all with the same value for the target attribute, return a single leaf node with that value;

(4) if R is empty, then return a single node with the value of the most frequent of the values of the target attribute that are found in records of S;
(5) Let A be the attribute with largest Gain(A,S) among attributes in R;
(6) Let {$a_j$ | j=1,2,…..,m} be the values of attribute A;
(7) Let {$S_j$ | j=1,2, .., m} be the subsets of S consisting respectively of records With value aj for A;
(8) Return a tree with root labeled D and arcs labeled $d_1$, $d_2$, .., dm going respectively to the

trees(HARC(R-{D},C,S1),HARC(R-{D},C,$S_2$), ....,HARC(R-{A}, C, Sm);
(9) Recursively apply HARC to subsets {$S_j$ | j=1,2, ..,m} until they are empty
(10) End

In the traditional classification approach single classification methods has been used like ARM or Decision Tree Classifier, whereas in this proposed method association rule and decision tree combined features have been used for the medical image classification.

## 5. RESULTS AND DISCUSSIONS

An experiment has been conducted on a CT scan brain image data set based on the proposed flow diagram as shown in Fig 1. The pre-diagnosed databases prepared by physicians are considered for decision making. Fig. 7 represents the original input image and Fig. 8 shows the result of histogram equalization of original image, which is used to reduce the different illumination conditions and noises at the scanning phase.

Median filterd technique has been used to find the edge feature in the CT scan brain image. Using this filter speckle noises are also removed. Fig 9 shows themedian filtered CT scan brain image. Morphological operation has been done before edge detection. Opening removes small objects from the foreground (usually taken as the dark pixels) of an image, placing them in the background.

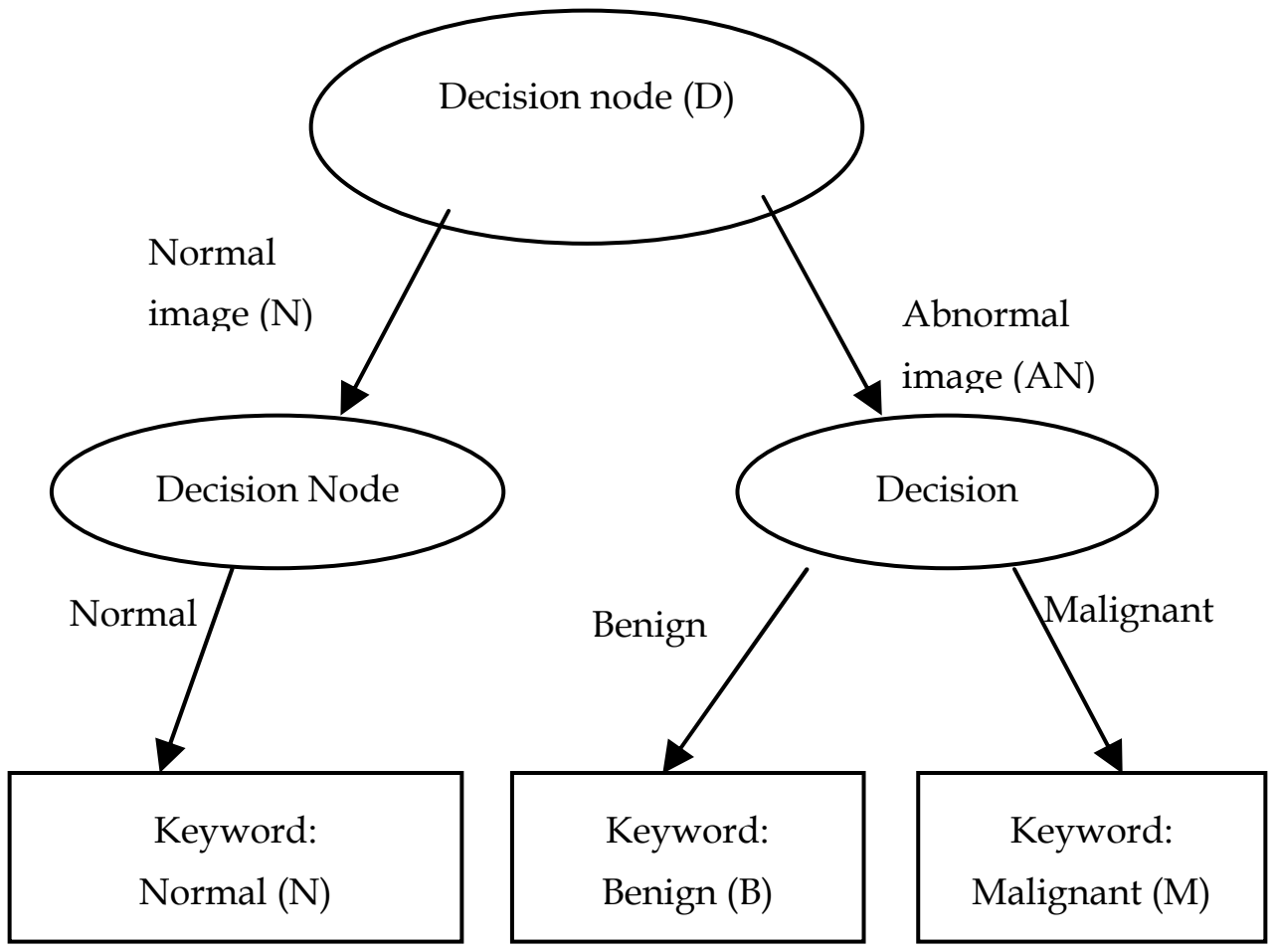

Fig 6.Classification of tumors cell using HARC

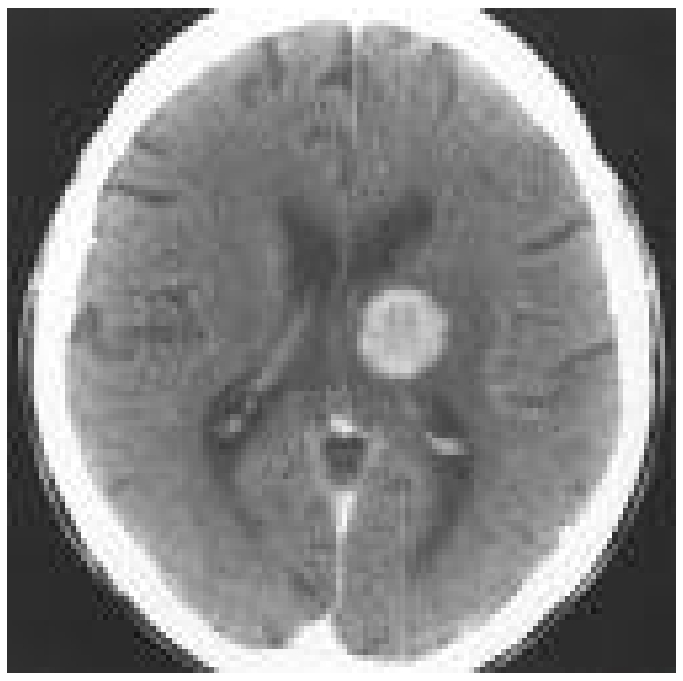

Fig. 7 Input CT scan brain image

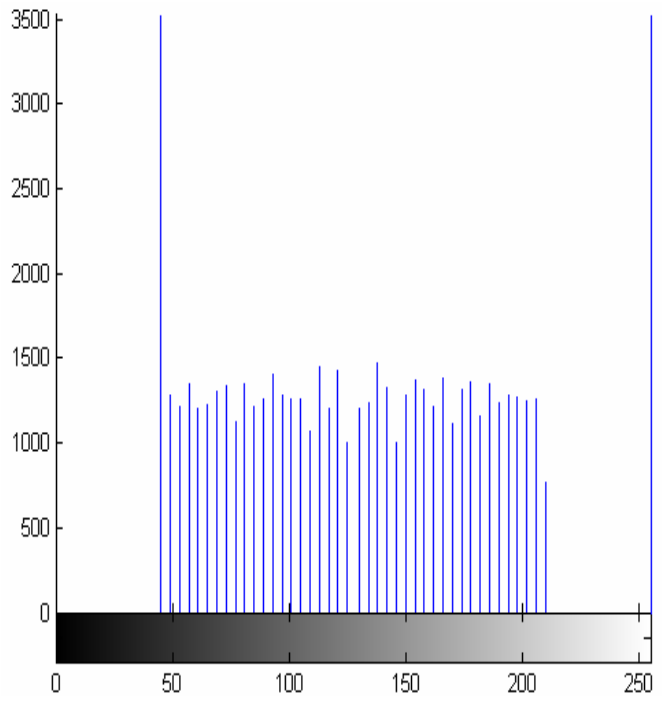

Fig. 8 Hitogram equalized image

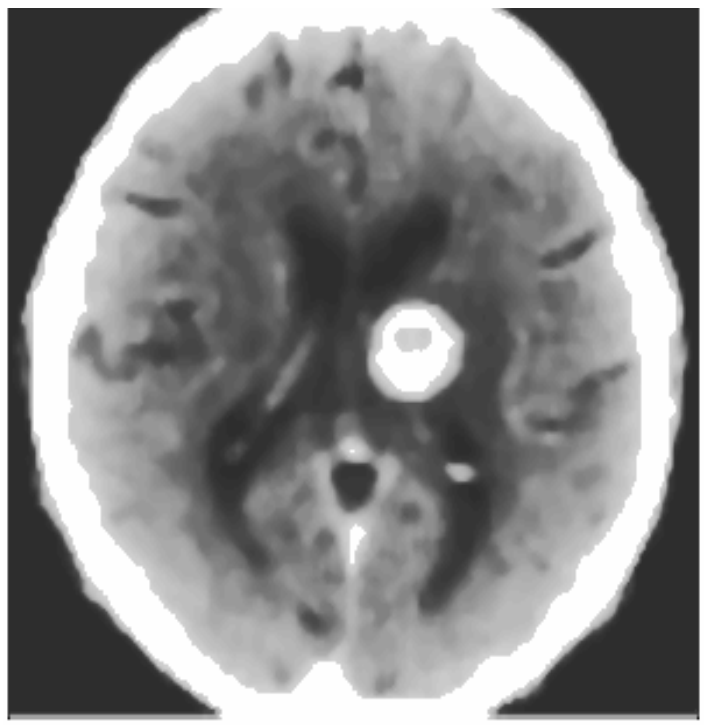

Fig. 9 Median filterd CT scan brain image

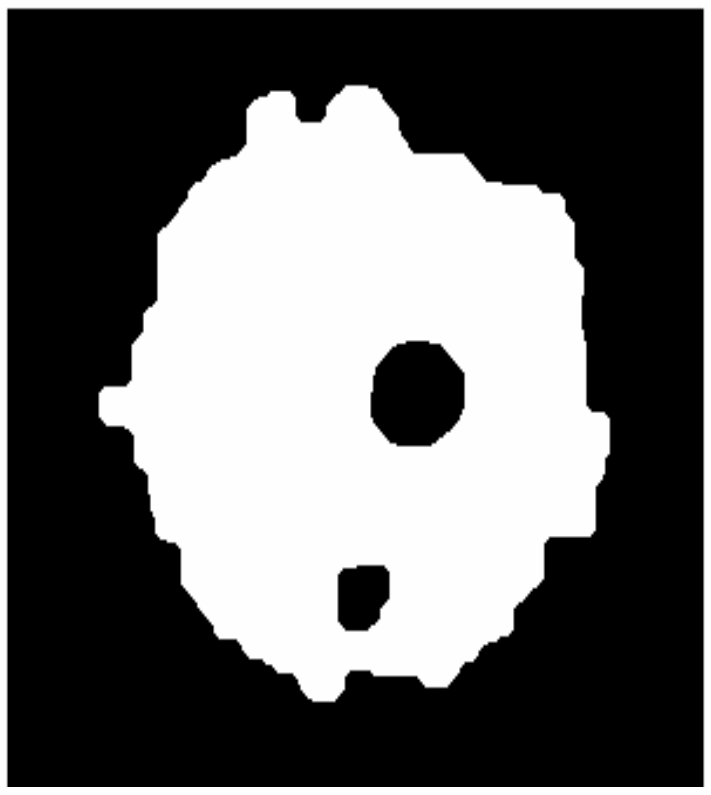

Fig. 10 Morphological opening image

This technique can also be used to find specific shapes in an image and also smooths the contour of an object. Morphological opened image is shown in Fig 10. For the object segregation, the edge features have been used. It has been done using canny edge detection. Fig 11 shows the segmented portion of an image. Fig 12 represents the constructed image by merging the segmented portion with the input of CT scan bain image. From the edge detected image the objects are extracted and it is shown in Fig 13 and Fig 14.

Texture features have been calculated for all segmented objects and stored in the transactional database.From the transactional database the association rules have been generated using the FP tree algorithm. The generated association rules are classified using the decision tree and it is shown in Fig .6.In this propsed method Hybrid Association Rule Classifier (HARC) have been generated, which is the conbined approach to classify the CT brain images into three categories namely normal, benign and malignatnt.

The results show that the proposed method can have better accuracy, sensitivity and specificity than the existing classification method. Depending on the tree build from the HARC algorithm the diagnosis can be made by both the physicians and the proposed system. The proposed system gives better results than the single classifier methods like C4.5 and Association Rule Classifier, it is decribed in the table 2 and 3 [26].

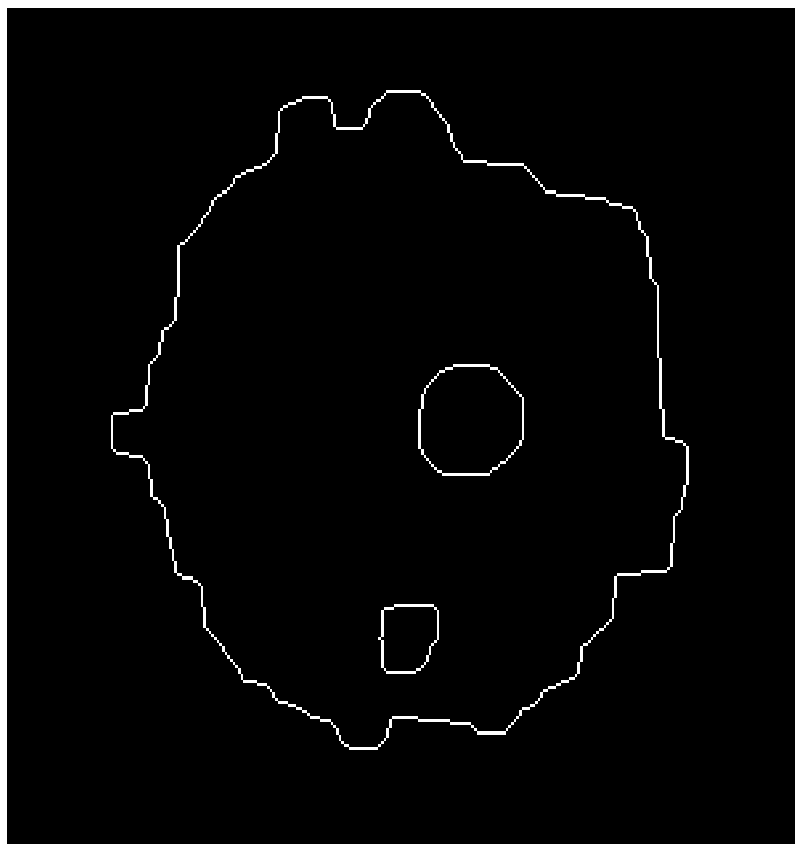

Fig. 11 Segmented output image

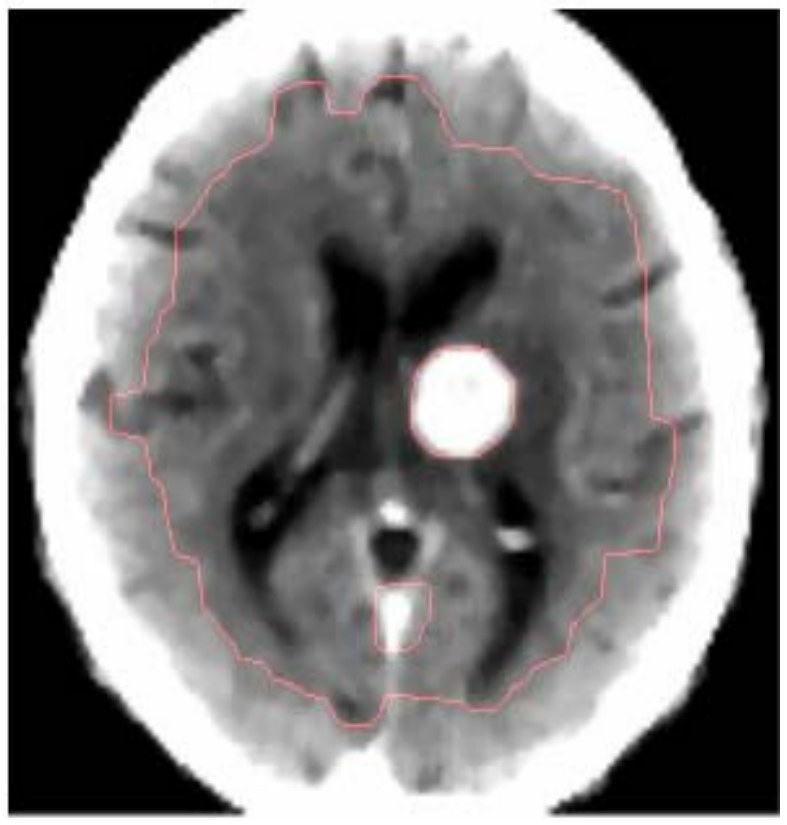

Fig. 12 Constructed image

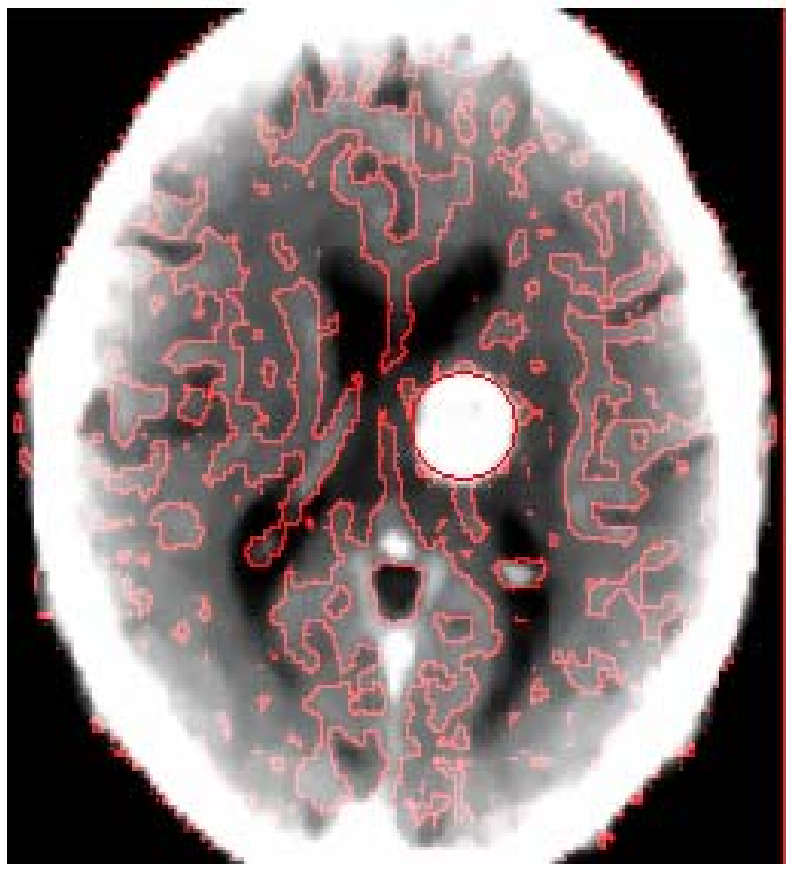

Fig. 13 Segregated image with more objects

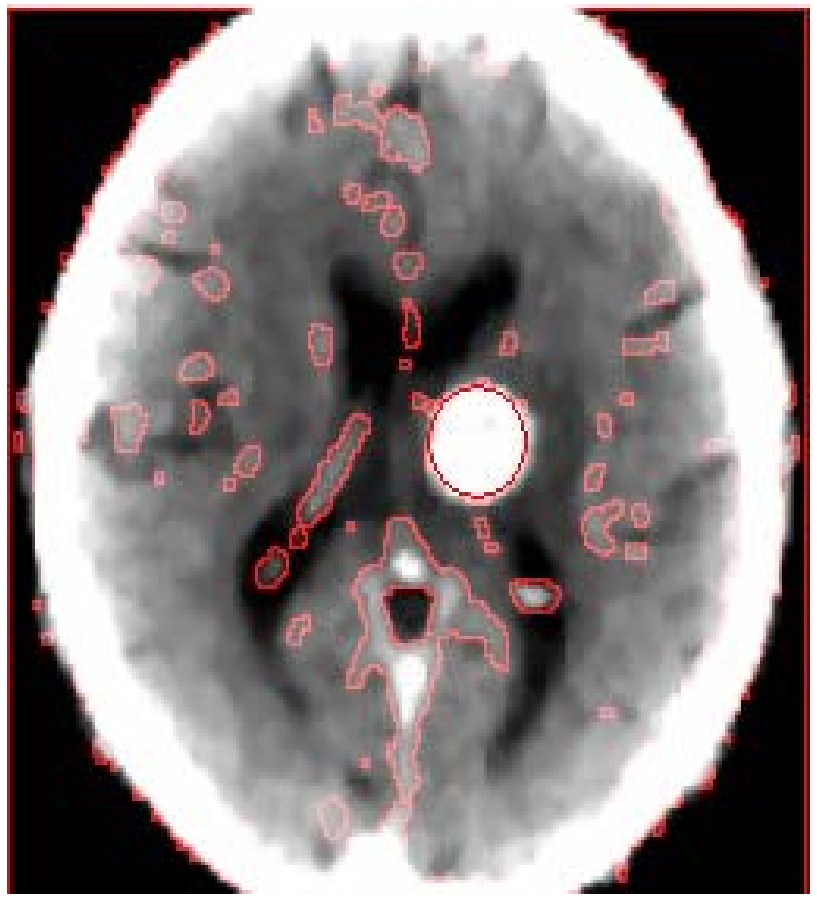

Fig. 14 Segregated image with less objects

The effectiveness of the proposed method has been estimated using the following measures:

Accuracy= (TP+TN)/ (TP+TN+FP+FN)

Sensitivity= TP/ (TP+FN)

Specificity= TN/ (TN+FP)

where, TP, TN, FP, and FN are the number of True Positive cases (abnormal cases correctly classified), the number of True Negatives (normal cases correctly classified), the number of False Positives (normal cases classified as abnormal), and the number of False Negatives (abnormal cases classified as normal) respectively. Accuracy is the proportion of correctly diagnosed cases from the total number of cases. Sensitivity measures the ability of the proposed method to identify abnormal cases. Specificity measures the ability of the method to identify normal cases. The value of minimum confidence is set to 97% and value of minimum support is set to 10%.

The features of the test images and the association rules have been generated using the threshold value=0.001. The results show that the proposed classifier gives higher values of sensitivity, specificity and accuracy such as 97%, 96% and 95% respectively. In order to validate the ob-tained results, the algorithmic approach has been compared with the well known classifier, C4.5 classifier and associative classifier [26], [27]. Table 2 and Table 3 represent the classification results and its performance with the existing classifiers. The proposed method gives better results as compared with the existing methods with respect to recall and precision value and it is shown in Fig. 15.

TABLE 2 CLASSIFICATION OF BRAIN TUMOR WITH ASSOCIATION RULE WITH DECISION TREE

| Categories | Physician | | | Association rule with decision tree classification | | |
|---|---|---|---|---|---|---|
| | Bengin* | Malignant* | Normal* | Bengin* | Malignant* | Normal* |
| Bengin* | TN 50 | FN 2 | TP 30 | TN 66 | FN 1 | TP 30 |
| Malignant* | FP 10 | TN 28 | TP 20 | FP 4 | TP 29 | TP 20 |
| Normal* | TP 5 | TP 5 | TP 10 | TP 5 | TP 5 | TP 10 |
| Total | 65 | 35 | 60 | 65 | 35 | 60 |

TABLE 3 COMPARISION OF PERFORMACE OF ALGORITHMS [26]

| Measure | Proposed Hybrid Method (HARC) | C4.5 | Associationrule |
|---|---|---|---|
| Sensitivity % | 97 | 84 | 95 |
| Accuracy % | 95 | 71 | 84 |
| Specificity % | 96 | 79 | 91 |

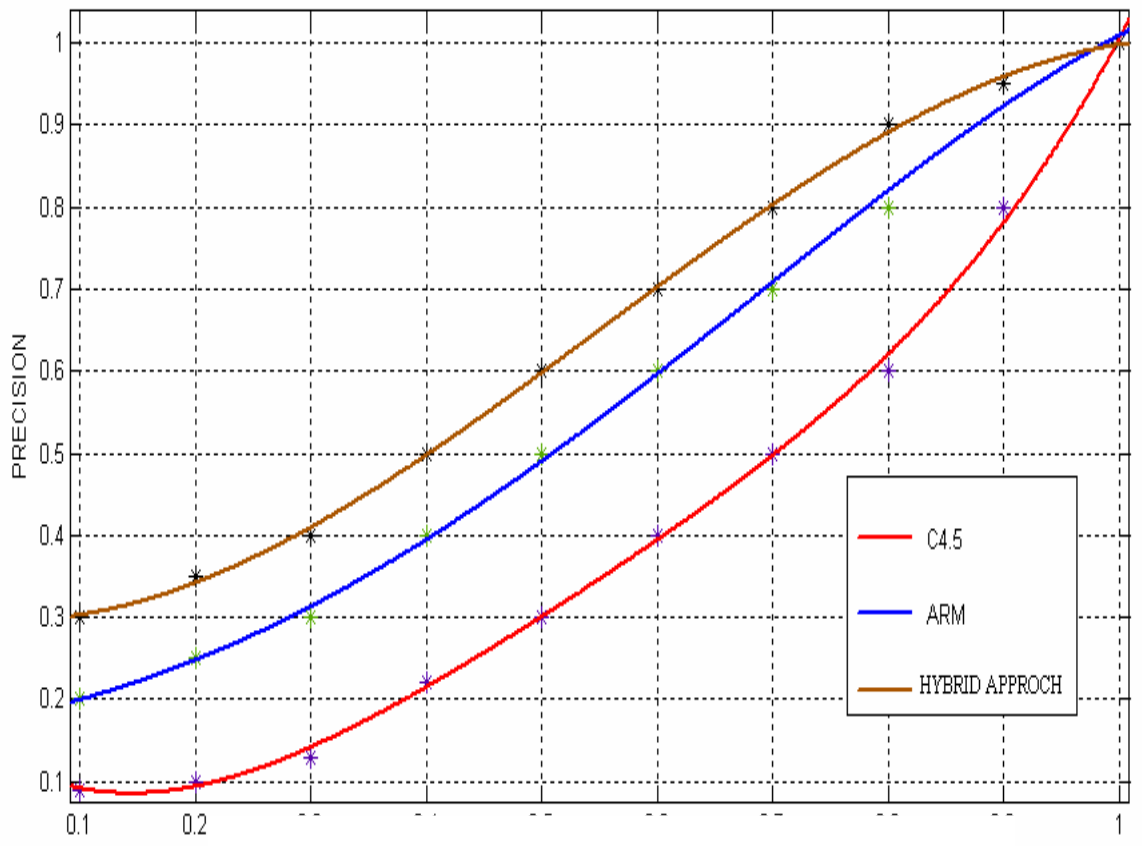

Fig. 15 Precision and recall graph

## 6. CONCLUSION

The hybrid image mining technique for brain tumor classification using association rule with decision tree method has been developed and performances evaluated. The median filtering techniques have efficiently reduces the speckle noises present in the CT scan brain images. The extracted objects using canny edge detection technique provides better results as compared to conventional method. The proposed hybrid approach of association rule mining and decision tree algorithm classifies the brain tumors cells in an efficient way. The proposed algorithm has been found to be performing well compared to the existing classifiers. The accuracy of 95% and sensitivity of 97% were found in classification of brain tumors. The developed brain tumor classification system is expected to provide valuable diagnosis techniques for the physicians.

## ACKNOWLEDGMENT

The authors would like to express their gratitude to Dr. D. Elangovan, Pandima CT scan centre, Dindigul for providing the necessary images for this study.

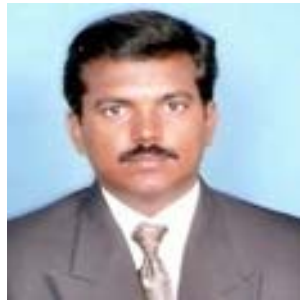

**P.Rajendran** obtained his MCA degree from Bharathidhasan University in 2000, ME Degree in Computer science and engineering from Anna University, Chennai, in 2005. He has started his teaching profession in the year 2000 in Vinayakamissions engineering college, salem. At present he is an Assistant Professor in department of computer science and engineering in K.S.Rangasamy college of Technology, Thiruchengode. . He has published 10 research papers in International and National Journals as well as conferences. He is a part time Ph.D research scalar in Anna University Chennai. His areas of interest are Data mining, Image mining and Image processing. He is a life member of ISTE.

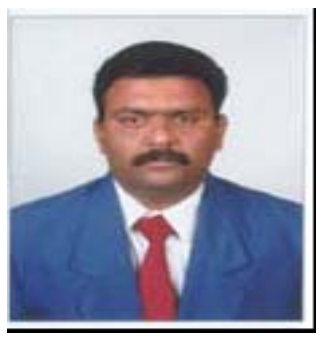

**Dr. M. Madheswaran** has obtained his Ph.D. degree in Electronics Engineering from Institute of Technology, Banaras Hindu University, Varanasi in 1999 and M.E degree in Microwave Engineering from Birla Institute of Technology, Ranchi, India. He has started his teaching profession in the year 1991 to serve his parent Institution Mohd. Sathak Engineering College, Kilakarai where he obtained his Bachelor Degree in ECE. He has served KSR college of Technology from 1999 to 2001 and PSNA College of Engineering and Technology, Dindigul from 2001 to 2006. He has been awarded Young Scientist Fellowship by the Tamil Nadu State Council for Science and Technology and Senior Research Fellowship by Council for Scientific and Industrial Research, New Delhi in the year 1994 and 1996 respectively. His research project entitled “Analysis and simulation of OEIC receivers for tera optical networks” has been funded by the SERC Division, Department of Science and Technology, Ministry of Information Technology under the Fast track proposal for Young Scientist in 2004. He has published 120 research papers in International and National Journals as well as conferences. He has been the IEEE student branch counselor at Mohamed Sathak Engineering College, Kilakarai during 1993-1998 and PSNA College of Engineering and Technology, Dindigul during 2003-2006. He has been awarded Best Citizen of India award in the year 2005 and his name is included in the Marquis Who's Who in Science and Engineering, 2006-2007 which distinguishes him as one of the leading professionals in the world. His field of interest includes semiconductor devices, microwave electronics, optoelectronics and signal processing. He is a member of IEEE, SPIE, IETE, ISTE, VLSI Society of India and Institution of Engineers (India).